%% file: main.tex
\documentclass{ecai} 

\usepackage{latexsym}
\usepackage{amssymb}
\usepackage{amsmath}
\usepackage{amsthm}
\usepackage{booktabs}
\usepackage{enumitem}
\usepackage{graphicx}
\usepackage{color}
\usepackage{algorithm}
\usepackage{algorithmic}
\usepackage{tabularx}
\usepackage{pifont}

\newcommand{\BibTeX}{B\kern-.05em{\sc i\kern-.025em b}\kern-.08em\TeX}

\begin{document}


\begin{frontmatter}


\paperid{123} 


\title{Ask Good Questions for Large Language Models}


\author[A,B]{\fnms{Qi}~\snm{Wu}}
\author[A,B]{\fnms{Zhongqi}~\snm{Lu}\thanks{Corresponding author. Email:zhongqi@cup.edu.cn}}

\address[A]{College of Artificial Intelligence, China University of Petroleum-Beijing, China}
\address[B]{Hainan Institute of China University of Petroleum (Beijing),  Sanya, Hainan, China}


\begin{abstract}
Recent advances in large language models (LLMs) have significantly improved the performance of dialog systems, yet current approaches often fail to provide accurate guidance of topic due to their inability to discern user confusion in related concepts. To address this, we introduce the Ask-Good-Question (AGQ) framework, which features an improved Concept-Enhanced Item Response Theory (CEIRT) model to better identify users' knowledge levels. Our contributions include applying the CEIRT model along with LLMs to directly generate guiding questions based on the inspiring text, greatly improving information retrieval efficiency during the question \& answer process.
Through comparisons with other baseline methods, our approach outperforms by significantly enhencing the users' information retrieval experiences.
\end{abstract}

\end{frontmatter}


\section{Introduction}

\begin{quote}
    \textit{A prudent question is one-half of wisdom.}\par\hfill --- Francis Bacon
\end{quote}

Asking the good questions is fundamental to effective information retrieval. Despite the powerful capabilities of large language models (LLMs) in understanding and reasoning \cite{openai2024openaio1card}, integrating these models into complex information retrieval tasks has demonstrated considerable potential \cite{guu2020realmretrievalaugmentedlanguagemodel, zhu2024largelanguagemodelsinformation}. The recent surge in AI-driven search engines, such as Perplexity and SearchGPT, underlines the growing demand for LLM-enhanced information retrieval. In this context, the ability to ask a good question becomes a powerful tool that can significantly enhance the effectiveness of these systems. By guiding the search process with strategically designed questions, users can shape the relevance and quality of the information retrieved. 
This capability positions question generation as a critical component in the next wave of LLM-driven information retrieval, where knowledge measure and targeted guidance is key to unlocking the full potential of LLMs in real-world applications. However, for all we know, the generation of guiding questions during the information retrieval process remains an underexplored area.

\begin{figure}[t]
    \centering
    \includegraphics[width=1.0\linewidth]{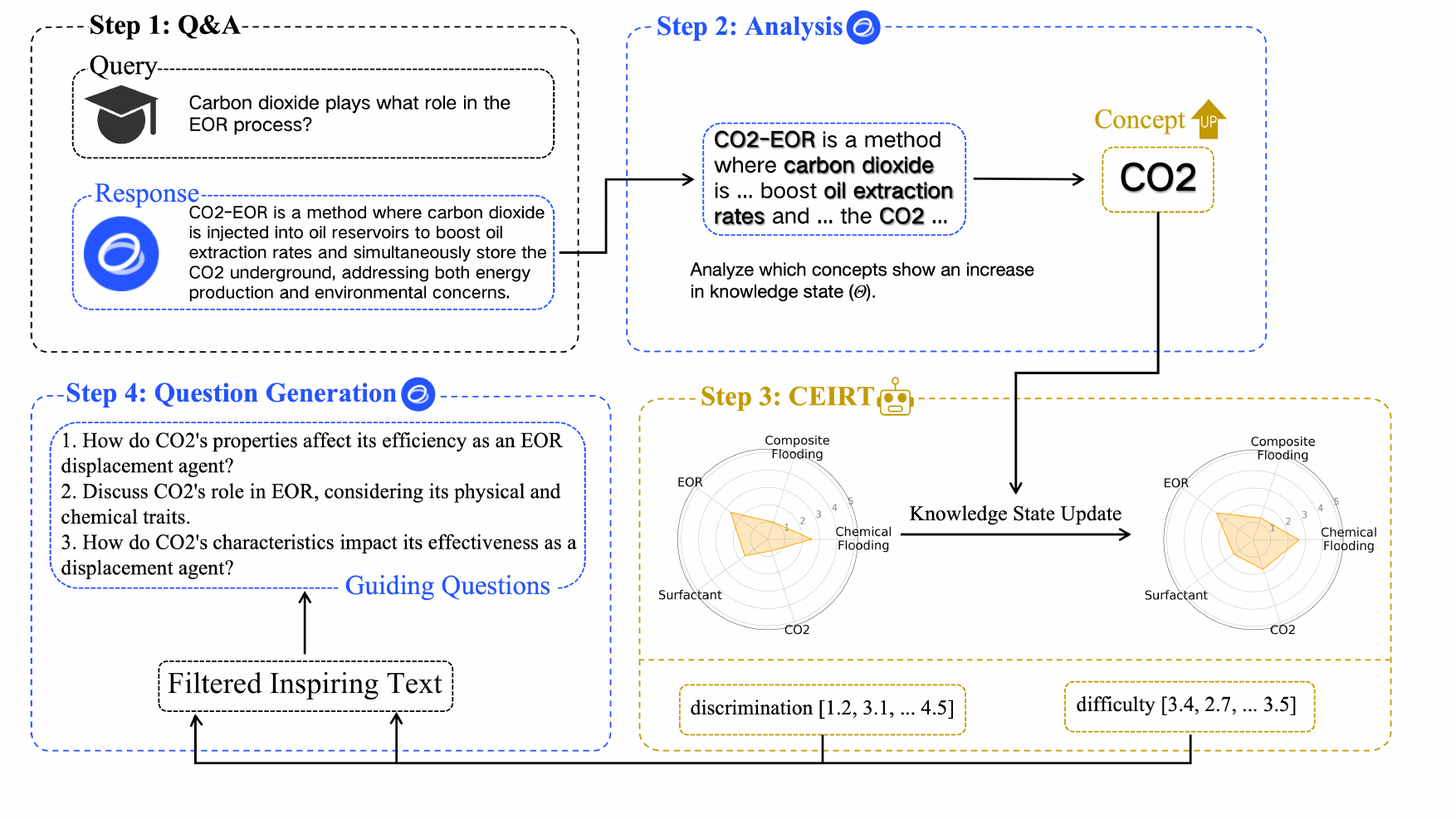}
    \caption{The diagram shows a single cycle of the Ask-Good-Question (AGQ) framework: processing user-LLM interactions, dynamically updating knowledge state vectors ($\boldsymbol{\theta}$), and using discrimination ($\boldsymbol{a}$) and difficulty ($\boldsymbol{b}$) parameters to filter inspiring texts for generating guiding questions that enhance information retrieval.}
    \label{fig:framework}
\end{figure}

Crafting good questions is often more difficult than answering them, as it requires a precise identification of knowledge gaps. 
This challenge becomes more significant for domain novices who face two difficulties: identifying their specific knowledge deficits and constructing questions that precisely target these information gaps. Both cognitive processes require conceptual frameworks that beginners typically have not yet developed.
Despite the strong contextual understanding of LLMs, these models struggle to identify user knowledge gaps in related concepts, limiting their ability to generate precise and relevant questions.
This challenge is further complicated by the poorly defined knowledge boundaries of LLMs in specialized domains. The limited understanding of these models' expertise creates uncertainty about their capability to generate questions that effectively guide information retrieval. Addressing these limitations requires an approach that dynamically assesses user knowledge states and formulates targeted questions based on contextual information, facilitating both immediate comprehension and deeper conceptual exploration.

To tackle these problems, we introduce the Ask-Good-Question (AGQ) framework. At the core of this framework lies the Concept-Enhanced Item Response Theory (CEIRT) model, which extends traditional psychometric approaches by incorporating conceptual dimensions into knowledge assessment. In our approach, knowledge points and questions encountered during information retrieval are modeled as assessable items. The CEIRT model represents key parameters as multi-dimensional vectors: the user's knowledge state ($\boldsymbol{\theta}$), item difficulty ($\boldsymbol{b}$), and item discrimination ($\boldsymbol{a}$). This vector-based approach allows for clear representation of both the user's understanding across various concepts and the characteristics of assessment items.
Levering these representations, the AGQ framework evaluates the user's conceptual understanding to identify knowledge gaps. This assessment enables the generation of guiding questions, the framework's primary output. These questions direct users to formulate effective queries and focus on critical areas, thereby improving information retrieval efficiency. Tailored to the user's knowledge state, these guiding questions provide adaptive support for knowledge acquisition, distinct from generic prompts.

The main contributions of this paper are as follows:
\begin{itemize}
    \item We develop a method that leverages LLMs to generate guiding questions by assessing the user's information needs and knowledge state, facilitating more efficient information retrieval in complex domains.
    
    \item We propose the CEIRT model to dynamically estimate the user's knowledge state ($\boldsymbol{\theta}$) across multiple concepts. The visualizations and experiments demonstrate that our CEIRT model provides reasonable measures of user understanding.
 
    \item We design a comprehensive evaluation framework to assess the effectiveness of guiding questions in improving information retrieval efficiency and knowledge acquisition.
\end{itemize}

\section{Related Work}

\subsection{Exploring the Knowledge Boundaries of LLMs}

Despite the impressive advancements of LLMs, their ability to explore and recognize their own knowledge boundaries remains limited. \cite{yin2023largelanguagemodelsknow} and \cite{ren2023investigatingfactualknowledgeboundary} investigate this aspect by designing question sets to challenge the models. On the other hand, works like \cite{li2024inferencetimeinterventionelicitingtruthful} and \cite{hernandez2024inspectingeditingknowledgerepresentations} focus on guiding the models during inference to increase the probability of generating correct answers. The R-Tuning method, introduced by \cite{zhang2024rtuninginstructinglargelanguage}, offers another perspective by constructing datasets and training LLMs to identify unanswerable questions within specific domains.

\subsection{Question-Answering Generation}

Question-Answer Generation involves automatically creating question-answer pairs from a given context. Early approaches often relied on rule-based and template methods, using linguistic patterns and predefined structures \cite{lindberg2013generating, curto2011exploring, labutov2015deep}. Neural sequence-to-sequence approaches marked a subsequent advance, works such as \cite{du2017learning,shakeri2020end} showing how to directly generate question-answer pairs from input text. Recently, research has begun to explore integrating Question-Answer Generation with LLMs, particularly through prompt engineering approaches \cite{zhang2024qwen}. 
We use dictionary-enhanced prompts to further ensure the accuracy and diversity of generated QA pairs. These pairs provide the structured knowledge for downstream tasks: modeling user knowledge states via CEIRT and generating guiding questions.

\subsection{Question Generation}

This task was initially proposed by \cite{vanderwende2007answering}. This foundational work spurred research into various methods and applications for QG. Subsequent advancements saw a shift towards neural sequence-to-sequence models and later, Transformer architectures \cite{vaswani2017attention}. Work by \cite{scialom2019self}, for instance, explored using Transformer architectures for answer-agnostic question generation. Their approach demonstrated the power of self-attention but focused on generating questions without requiring a predefined answer. Research from \cite{back2021learning} introduced a pre-training method aimed at generating contextually rich questions by learning to recover answer-containing sentences. This highlights efforts to improve question quality, often targeting downstream reading comprehension tasks. Some work, such as \cite{uto2023difficulty}, focused on controllability, using Item Response Theory (IRT) \cite{lord2012applications} to manage the difficulty of generated question-answer pairs. While valuable for creating assessments with specific difficulty levels, this approach differs from dynamically guiding users based on evolving knowledge states.
While much QG research focuses on applications like reading comprehension assessment or data augmentation \cite{back2021learning, uto2023difficulty}, our AGQ framework employs QG specifically to generate guiding questions that enhance information retrieval efficiency and improve user knowledge states. Leveraging CEIRT model, AGQ tailors these questions based on dynamically assessed knowledge gaps ($\boldsymbol{\theta}$) and item characteristics ($\boldsymbol{b}$), rather than generating general questions.

\subsection{IRT-Based Adaptive Assessment and Diagnosis}

Computerized Adaptive Testing (CAT), based on Item Response Theory (IRT), dynamically selects items to efficiently estimate an examinee's ability level \cite{weiss1984application}. The field has since been enhanced with techniques for content balancing and item exposure control \cite{van1998model}, extensions to multidimensional models (MIRT) for assessing multiple traits \cite{segall1996multidimensional}, and refined item selection strategies \cite{chang1999stratified}. However, the primary goal of these IRT/MIRT models in CAT is to yield a stable, final ability estimate ($\theta$). In contrast, our CEIRT model dynamically tracks a user's knowledge state ($\boldsymbol{\theta}$) throughout an interaction to *inform guidance* rather than produce a *terminal assessment*.

Cognitive Diagnostic Models (CDMs) also leverage IRT principles but focus on providing fine-grained profiles of skill mastery, often using a Q-matrix to map items to skills \cite{tatsuoka1983rule}. While advanced models like G-DINA can represent complex skill relationships \cite{de2011generalized}, CDMs typically diagnose mastery of discrete skills for summative feedback \cite{tatsuoka1983rule, de2011generalized}. Conversely, the CEIRT model utilizes a continuous multidimensional vector ($\boldsymbol{\theta}$) to represent a user's degree of understanding, updating it dynamically to serve as direct input for generating tailored guiding questions.

In summary, established IRT applications focus on assessment (CAT) and diagnosis (CDM). In contrast, the CEIRT model represents conceptual understanding via a continuous, dynamic, multidimensional state vector ($\boldsymbol{\theta}$). The primary contribution of the AGQ framework is the integration of this CEIRT model with LLMs to generate adaptive guiding questions based on $\boldsymbol{\theta}$ within interactive information retrieval dialogues.

\begin{algorithm*}[ht]
    \caption{Ask-Good-Question Framework}
    \label{alg:AGQ}
    \begin{algorithmic}[0]
        \REQUIRE User's query $q$, Tutor prompt $P_T$
        \REQUIRE LLM $f(\cdot)$, CEIRT model $g(\cdot)$, Filtering model $h(\cdot)$
        \REQUIRE Guiding question generation prompts $P_{QGlow}$ and $P_{QGhigh}$ for low and high knowledge states
        \REQUIRE Lower bound $\epsilon$, Concept set $C$
        \WHILE{``exit'' not in $q$}
        \STATE $R \leftarrow f(q, P_T) \hfill \triangleright \text{Generate response }R\text{ for user's query}$
        \STATE $C \leftarrow f(R) \hfill \triangleright \text{Identify concepts with knowledge state changes}$ 
        \STATE $\mathbf{b}, \mathbf{a}, \boldsymbol{\theta} \leftarrow g(C) \hfill \triangleright \text{Update difficulty }\mathbf{b}\text{, discrimination }\mathbf{a}\text{ and knowledge states }\boldsymbol{\theta}$
        \STATE $\theta_{low} \leftarrow \text{None} \hfill \triangleright \text{Initialize low knowledge state flag}$
        \FOR{$\theta_j$ in $\boldsymbol{\theta}$}
            \IF{$\theta_j \leq \epsilon$}
                \STATE $\theta_{low} \leftarrow \theta_j \hfill \triangleright \text{Select concept }j\text{ with low knowledge state, store value for potential use}$
                \STATE \textbf{break} \hfill $\triangleright$ \text{Prioritize lowest knowledge state, exit loop}
            \ENDIF
        \ENDFOR
        \IF{$\theta_{low}$ is not None} 
            \STATE $T \leftarrow h(C,\mathbf{a},\mathbf{b},\theta_{low}) \hfill \triangleright \text{Filter text based on the identified low knowledge state concept}$
            \STATE $Q \leftarrow f(P_{QGlow}, T) \hfill \triangleright \text{Generate understanding-biased questions}$
        \ELSE 
            \STATE $T \leftarrow h(C,\mathbf{a},\mathbf{b},\boldsymbol{\theta}) \hfill \triangleright \text{Filter relevant inspiring text }T \text{ based on overall state}$
            \STATE $Q \leftarrow f(P_{QGhigh}, T) \hfill \triangleright \text{Generate application-biased questions}$
        \ENDIF
        \ENDWHILE
    \end{algorithmic}
\end{algorithm*}

\section{Methodology}

In this section, we detail the runtime operation of the AGQ framework, which enhances information retrieval through dynamic, CEIRT-based knowledge assessment and adaptive guiding question generation. The framework utilizes LLMs and operates on domain-specific knowledge resources. We describe each component below.

\subsection{CEIRT Model}

Traditional Item Response Theory (IRT) models typically assume a one-dimensional ability parameter, insufficient for updating knowledge states in information retrieval. To address this, we propose the Concept-Enhanced Item Response Theory (CEIRT) model, which extends the Multidimensional Item Response Theory (MIRT) framework, adapting the structure of the 2-Parameter Logistic (2PL) model. CEIRT uses vector representations for key parameters: the user's knowledge state across $K$ concepts $\boldsymbol{\theta} \in \mathbb{R}^K$, the difficulty parameter $\boldsymbol{b} \in \mathbb{R}^K$ for question $i$, and the discrimination parameter $\boldsymbol{a} \in \mathbb{R}^K$ for question $i$, indicating how well it differentiates between knowledge states.

The knowledge state vector $\boldsymbol{\theta}$ is implemented using an embedding layer, offering a flexible representation of user understanding in concepts. As a latent representation learned from data, the embedding provides an estimation of the user's underlying knowledge state, rather than a direct psychometric measurement. More importantly, this knowledge representation is dynamic. 
As depicted in Algorithm \ref{alg:AGQ}, parameters $\boldsymbol{\theta}$, are adjusted through gradient-based optimization using user-LLM interaction data. This optimization process refines the knowledge state estimates by minimizing the Binary Cross-Entropy (BCE) loss between the model's predicted probabilities of correct responses and the actual user outcomes ($y_i$). This allows $\boldsymbol{\theta}$ to continuously evolve and effectively capture the relative changes in the user's understanding during interaction. Building upon the logistic function characteristic of the 2PL model, the probability $p_i$ that a user provides a correct response to question $i$ in our multidimensional context is calculated as:
\begin{equation}
p_i = \frac{1}{1 + \exp(-\sum_j(a_i\theta_j - b_i))}
\label{eq:prob_correct}
\end{equation}
This formulation captures how multiple conceptual abilities ($\theta_j$) contribute, weighted by the question's discrimination ($a_i$), relative to its difficulty ($b_i$). It is important to note that within this model, a higher value for a specific component $\theta_j$ in the knowledge state vector $\boldsymbol{\theta}$ corresponds to a higher understanding of the user regarding concept $j$.

\begin{figure}[t]
    \centering
    \includegraphics[width=0.6\columnwidth]{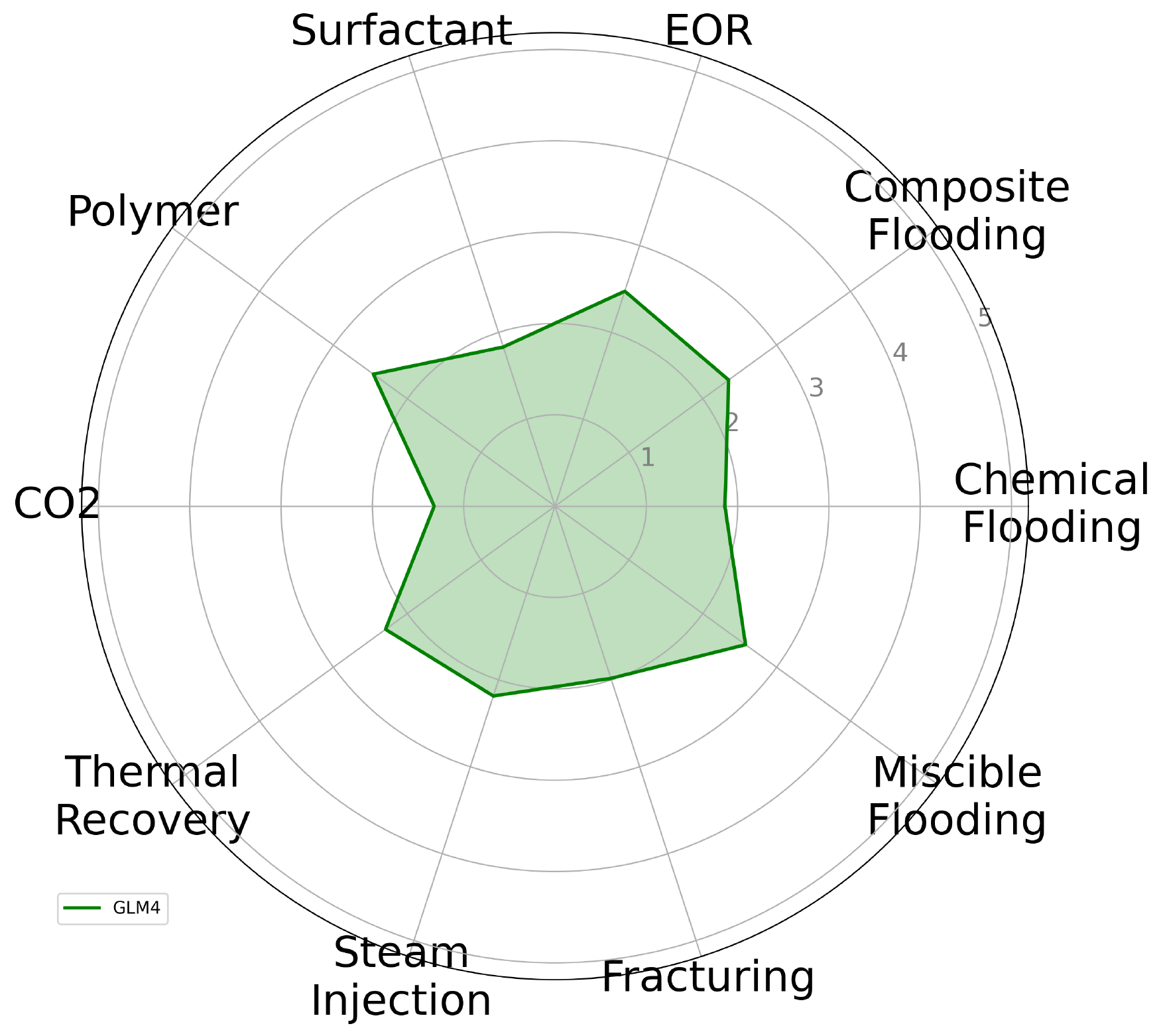}
    \caption{Radar chart of LLM (ChatGLM4-9B-Chat example) knowledge state ($\boldsymbol{\theta}$ from CEIRT) in the petroleum domain. Low $\boldsymbol{\theta}$ values across concepts indicate insufficient expertise for independent guiding question generation.}
    \label{fig:radar-glm4}
\end{figure}

\subsection{Dataset Construction}
\label{dataset-construction}

At the outset of our research, we conducted a preliminary assessment of the LLM's inherent knowledge within specialized domains. This assessment using CEIRT revealed significant limitations in the model's conceptual understanding of vertical fields. The results, visualized in Figure \ref{fig:radar-glm4}, demonstrated clear gaps in the LLM's knowledge base (using ChatGLM4-9B as example).

To address this limitation and provide the necessary structured knowledge for the AGQ framework, we implemented a multi-stage dataset generation pipeline leveraging LLM capabilities.
Given a corpus of domain-specific documents $D = \{d_1, ..., d_N\}$, we first prompt an LLM to automatically identify a set of key concepts $C = \{c_1, ..., c_K\}$ by analyzing the document content, for instance, from document abstracts and keywords. 
Sentence extraction for each concept $c_k \in C$ was then performed via prompting, identifying relevant sentences $S_k = \{s_{k,1}, s_{k,2}, ...\}$ from the corpus $D$ related to $c_k$: $S_k = \text{LLM}_{\text{extract}}(D, c_k)$. These extracted sentences were organized into a dictionary $\text{Dict} = \{c_k: S_k\}_{k=1}^K$, mapping each concept to its associated example sentences.
For each sentence $s_{k,m} \in \text{Dict}_{k}$, a corresponding question-answer pair $(q_{k,m}, a_{k,m})$ was generated, where the question targets concept $c_k$ within the sentence: $(q_{k,m}, a_{k,m}) = \text{LLM}_{\text{QG}}(\text{Dict}_{k,m})$. 
This resulted in a preliminary set of QA pairs associated with specific concepts. Finally, all generated QA pairs underwent manual verification by domain experts to ensure accuracy, relevance, and clarity, yielding the final dataset used by the AGQ framework. 

\subsection{User-LLM Interaction}
User interaction commences with a query related to their information need. The framework utilizes an LLM, guided by a structured prompt, to generate a relevant response $R$. The prompt includes the user's initial knowledge state vector $\boldsymbol{\theta}$, converted into a textual description of proficiency levels, to contextualize the response generation; further details are provided in the Appendix.
Following the generation of response $R$, the framework dynamically updates the user's estimated knowledge state $\boldsymbol{\theta}$ to reflect potential learning from the interaction. This process involves two primary steps integrated within the AGQ workflow (Algorithm \ref{alg:AGQ}):

First, an LLM-based analysis module $f(R)$ examines the content of the response $R$. This module leverages the LLM's contextual understanding to identify a subset of key concepts $C \subseteq C_{all}$ where $C_{all}$ is the set of all predefined concepts. This step infers the conceptual areas that the user may have engaged with during the information retrieval process.

Second, the inferred conceptual needs to be translated into a format suitable for updating the CEIRT model parameters. The CEIRT model estimates knowledge states based on user interaction history, typically represented as responses to assessment items. To incorporate the learning inferred from the unstructured response $R$, we employ a simulated evidence generation. For each concept $c \in C$, the user's interaction history is augmented by simulating a correct response to a virtual assessment item associated with concept $c$. This effectively creates proxy evidence reflecting the inferred knowledge gain.

This augmented interaction history, incorporating the simulated correct responses for concepts in $C$, serves as new input for the CEIRT model's update step $g(C)$. During this step, $\boldsymbol{\theta}$ is refined by performing further optimization iterations (e.g., a few epochs of gradient descent using an Adam optimizer) to minimize the BCE loss over the cumulative interaction history.
The resulting updated knowledge state vector $\boldsymbol{\theta}$ provides an estimate of the user's current understanding across all concepts, forming the basis for the subsequent question generation phase.

\subsection{Inspiring Text}
Effective guiding question generation by the LLM relies on providing appropriate contextual input, termed 'Inspiring Text'. The selection of this text is guided by the principle of optimal challenge, aiming to match the text's cognitive demand with the user's current knowledge state to foster information retrieval.

To implement this principle, we define a suitability score $S(t,j)$ for a candidate text fragment $t$. This text $t$ is associated with concept $j$ and possesses a quantified difficulty parameter $b_i$, derived from associated assessment data within the structured knowledge corpus mentioned in Section \ref{dataset-construction}. The score is calculated relative to the user's knowledge state $\theta_j$ in concept $j$:
\begin{equation} \label{eq:suitability_score}
S(t,j) = \exp(-(|{\theta_j - b_i}|-1)^2)
\end{equation}
This scoring function peaks when the absolute difference between the user's knowledge state and the text's difficulty parameter, $|{\theta_j - b_i}|$, equals 1. A perfect match ($|{\theta_j - b_i}|=0$) might offer insufficient cognitive challenge, while a large discrepancy could render the text incomprehensible; thus, a slight deviation is preferred to promote information retrieval.

Empirical validation for this peak value comes from our ablation study (Section \ref{ablation}), which demonstrates that user knowledge gain is maximized when $|{\theta_j - b_i}|$ is approximately 1.

For each target concept requiring a guiding question, the framework selects the text fragment(s) with the highest suitability score $S(t,j)$ to serve as the Inspiring Text. This selected text is then provided as contextual input to the LLM, enabling the generation of context-aware, user-adaptive, and appropriately challenging guiding questions.

Based on the updated knowledge state $\boldsymbol{\theta}$, the AGQ framework employs an adaptive strategy for generating guiding questions, as outlined in Algorithm \ref{alg:AGQ}. If a concept $j$ is identified with a knowledge state $\theta_j$ below a predefined threshold $\epsilon$, indicating a need for foundational understanding, the LLM utilizes a specific prompt ($P_{QGlow}$), designed to elicit foundational questions (e.g., 'What is ...?'), with the selected Inspiring Text. Conversely, if all relevant concepts exceed the threshold, a different prompt ($P_{QGhigh}$), crafted to encourage application-focused questions (e.g., 'How can ... be applied to ...?'), is used to generate application-biased questions, focusing on practical application and deeper exploration.

\subsection{Relevance and Quality Control of Guiding Questions}

To ensure that guiding questions generated by the AGQ framework are relevant to the user's information retrieval needs and maintain a suitable standard, a quality assessment mechanism is employed. This mechanism evaluates potential guiding questions based on several key attributes, allowing for the identification and filtering of questions that may be poorly aligned or low-quality.
The assessment incorporates the following metrics:
First, information gap alignment is considered. For a concept $j$, this is quantified as:
\begin{equation}
    AlignScore = 1 - \theta_j
    \label{eq:align_score}
\end{equation}
where $\theta_j$ represents the user's estimated knowledge state in concept $j$. A higher $\textit{AlignScore}$ indicates that the question targets an area where the user's current knowledge state suggests a potential information gap, making the question more pertinent to their retrieval objectives.

Second, conceptual specificity measures the focus of the question. Using mutual information between a question $q$ and its target concept $c$:
\begin{equation}
    MI(q,c) = \sum_{q,c} p(q,c) \log \frac{p(q,c)}{p(q)p(c)}
    \label{eq:mutual_info_qc}
\end{equation}
where $p(q,c)$ is the joint probability and $p(q)$, $p(c)$ are the marginal probabilities. This metric quantifies the strength of association, ensuring the question is specific to the intended concept rather than overly general.

Third, linguistic complexity evaluates the question's structure. Representing the token count of question $q$ as $len(q)$, the complexity index is calculated via standardization and sigmoid transformation:
\begin{equation}
    ComplexityIndex = \sigma\left(\frac{len(q) - \mu}{\sigma}\right)
    \label{eq:complexity_index}
\end{equation}
where $\sigma(x) = (1 + e^{-x})^{-1}$ is the sigmoid function, and $\mu$, $\sigma$ are the mean and standard deviation of token counts in the dataset. This index helps assess if the question's formulation is appropriately complex for clear guidance, avoiding overly simplistic or convoluted phrasing.

These metrics are combined into a final quality score:
\begin{equation}
    QualityScore = \alpha \cdot AlignScore + \beta \cdot MI(q,c) + \gamma \cdot ComplexityIndex
    \label{eq:quality_score}
\end{equation}
where $\alpha, \beta, \gamma$ are empirically determined weight coefficients satisfying $\alpha + \beta + \gamma = 1$. This $\textit{QualityScore}$ provides a quantitative basis for assessing each potential guiding question. It is used for analysis or as a criterion for filtering out questions that fall below a predetermined quality threshold, thus improving the overall relevance and effectiveness of the guidance provided within the information retrieval process.

\section{Experiment Setup}
To validate the effectiveness of AGQ framework, we designed and conducted a series of experiments. The primary goal was to evaluate the performance of AGQ framework across different scenarios. It is important to note that AGQ framework does not have specific requirements for LLMs and can be applied to both open-source and proprietary models. For these experiments, we selected ChatGLM4-9B as the representative model, which demonstrates strong performance in NLP tasks\cite{glm2024chatglmfamilylargelanguage}.

\subsection{Dataset}
The experiments utilized the EOR-QA dataset, a custom dataset built for AGQ in the Enhanced Oil Recovery (EOR) domain. This dataset links key EOR concepts to relevant contextual sentences and paragraphs extracted from domain literature. Based on this structured context, EOR-QA comprises over 3,100 question-answer pairs, designed to cover different cognitive levels, from foundational principles to practical applications. As elaborated in Section \ref{dataset-construction}, all content underwent verification by domain experts to ensure accuracy and relevance. The EOR-QA dataset serves a dual purpose: it provides the foundational data for the CEIRT model to estimate and update user knowledge states ($\boldsymbol{\theta}$), and it acts as the source repository for selecting Inspiring Text used in guiding question generation.

The decision to construct a custom dataset stemmed from a survey of existing public resources, which revealed a lack of datasets adequately addressing the needs of our research in the EOR domain. While numerous general/scientific QA datasets exist, they typically lack the required domain-specific depth and concept coverage that are crucial for the AGQ framework's user modeling and adaptive guidance mechanisms. Furthermore, our custom dataset mitigates the risk of data contamination inherent in public benchmarks, ensuring a rigorous evaluation of the model's performance on genuinely unseen domain knowledge. Therefore, the creation of the EOR-QA dataset was a necessary prerequisite to effectively implement and reliably evaluate the AGQ framework within this specialized domain.

\subsection{Baselines}
We compared the AGQ framework against several baseline methods to evaluate its effectiveness. The following approaches were included in our experimental comparison:

\begin{itemize}
    \item \textbf{Zero-shot Question Generation:} 
    We directly prompted the LLM to generate guiding questions without providing any examples. This method relies solely on the LLM's pre-trained knowledge and capabilities to generate relevant questions based on the user's query.
    \item \textbf{CoT Prompts with Handcrafted Examples:}
    This method incorporates Chain-of-Thought (CoT) prompting with manually constructed examples derived from the EOR-QA dataset to assist the LLM in guiding question generation, as CoT has demonstrated effectiveness in complex reasoning tasks requiring step-by-step thinking.
    \item \textbf{Human Expert:}
    We also consider the guiding questions generated by human experts in the petroleum field. These experts, with their deep domain knowledge, create questions that are specifically tailored to identify and address gaps in the user's understanding. This method serves as a gold standard against which the performance of the LLM-generated questions can be measured.
\end{itemize}

\begin{figure}[ht]
    \centering
    \includegraphics[width=1.0\columnwidth]{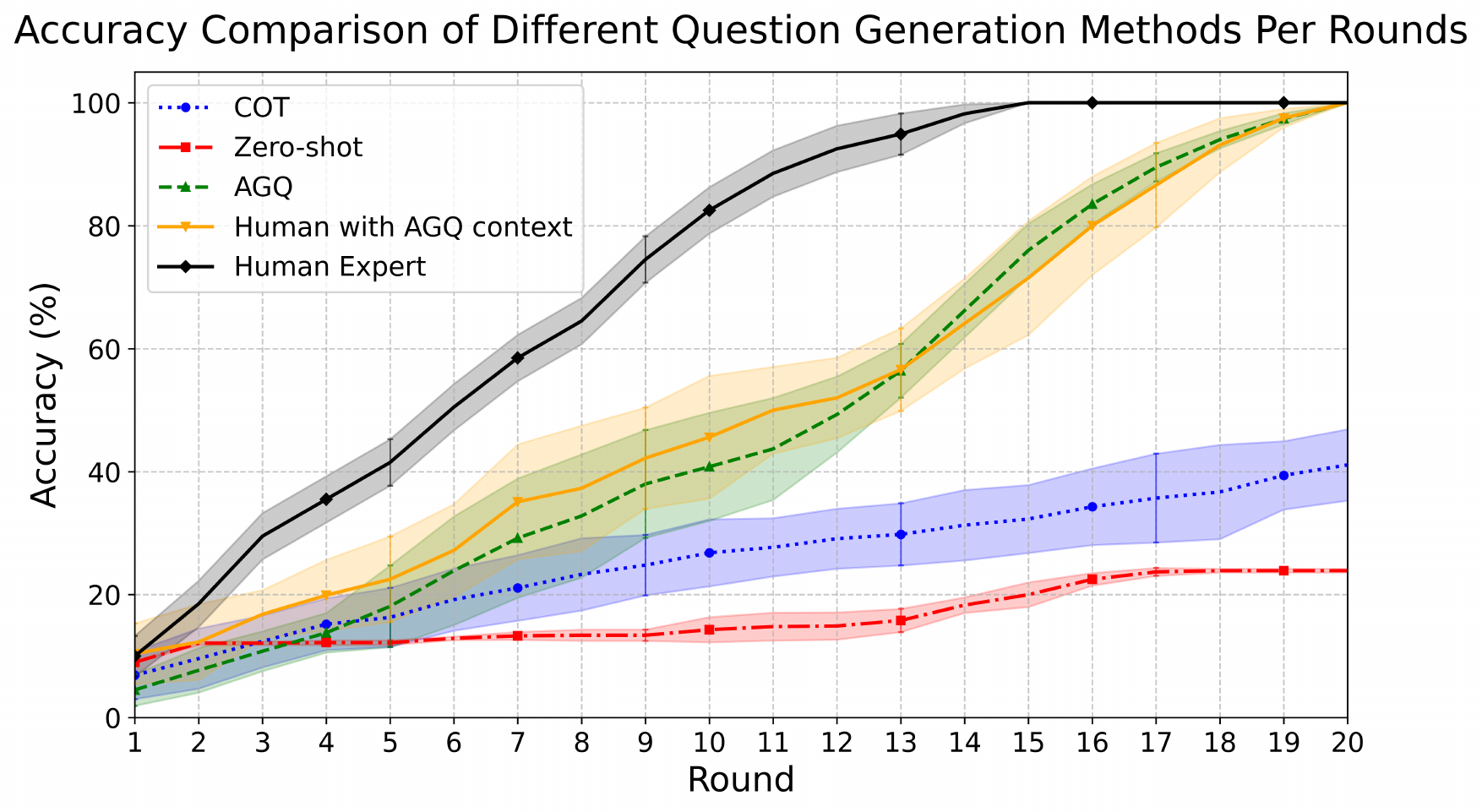}
    \caption{Accuracy comparison of different guiding question generation methods over dialogue rounds. The AGQ method demonstrates performance significantly exceeding CoT and Zero-shot approaches, closely approaching the effectiveness of Human Experts. The 'Human with AGQ context' group serves to validate the evaluation methodology.}
    \label{fig:baselines' accuracy}
\end{figure}

\begin{figure}[ht]
    \centering
    \includegraphics[width=1.0\columnwidth]{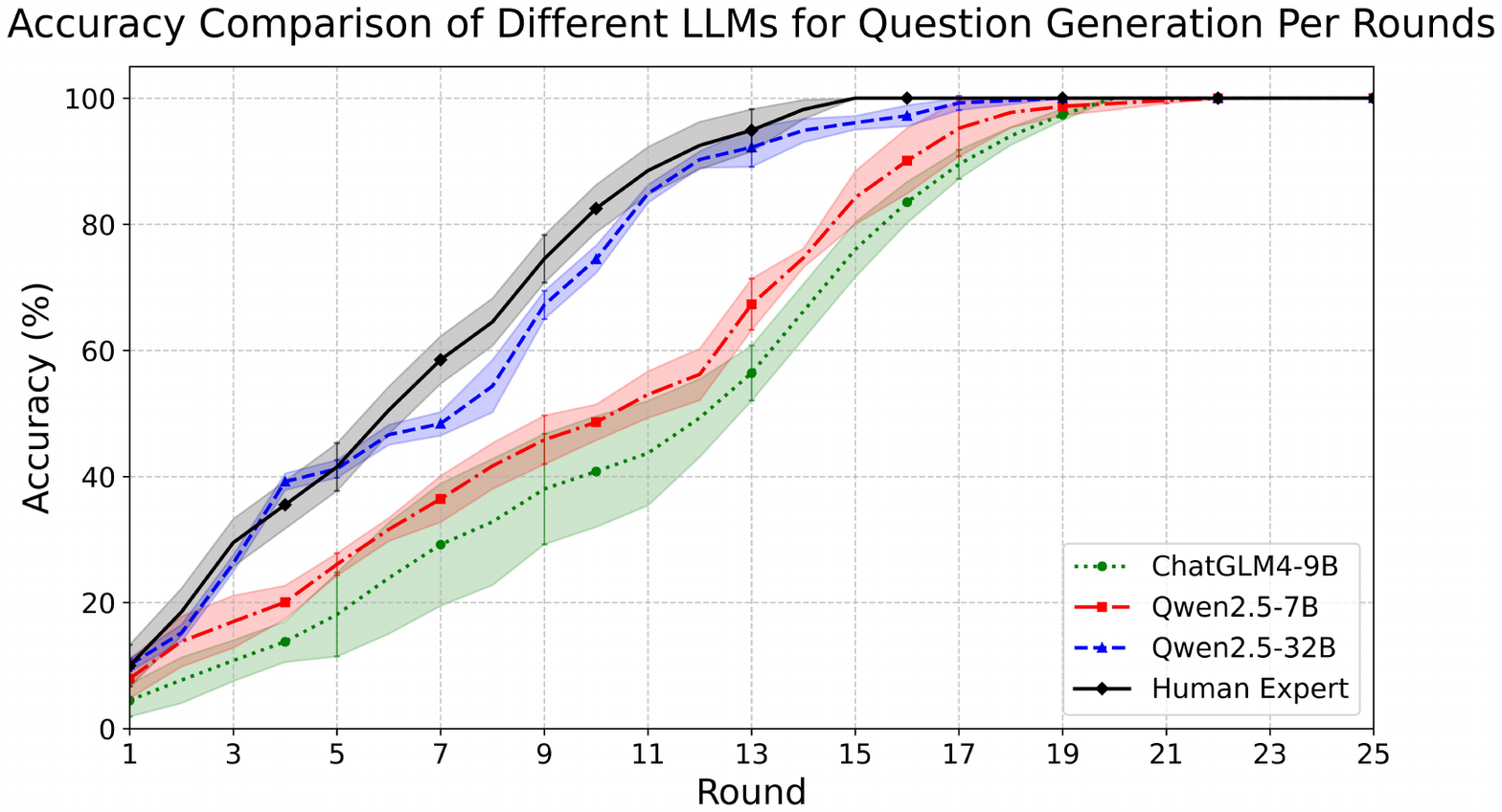} 
    \caption{Accuracy comparison of guiding question generation using the AGQ framework with different LLMs (ChatGLM4-9B, Qwen2.5-7B, Qwen2.5-32B) and Human Expert over dialogue rounds.}
    \label{fig:diff_llm_accuracy} 
\end{figure}

\begin{figure*}[ht!]
    \centering
    \includegraphics[width=0.3\linewidth]{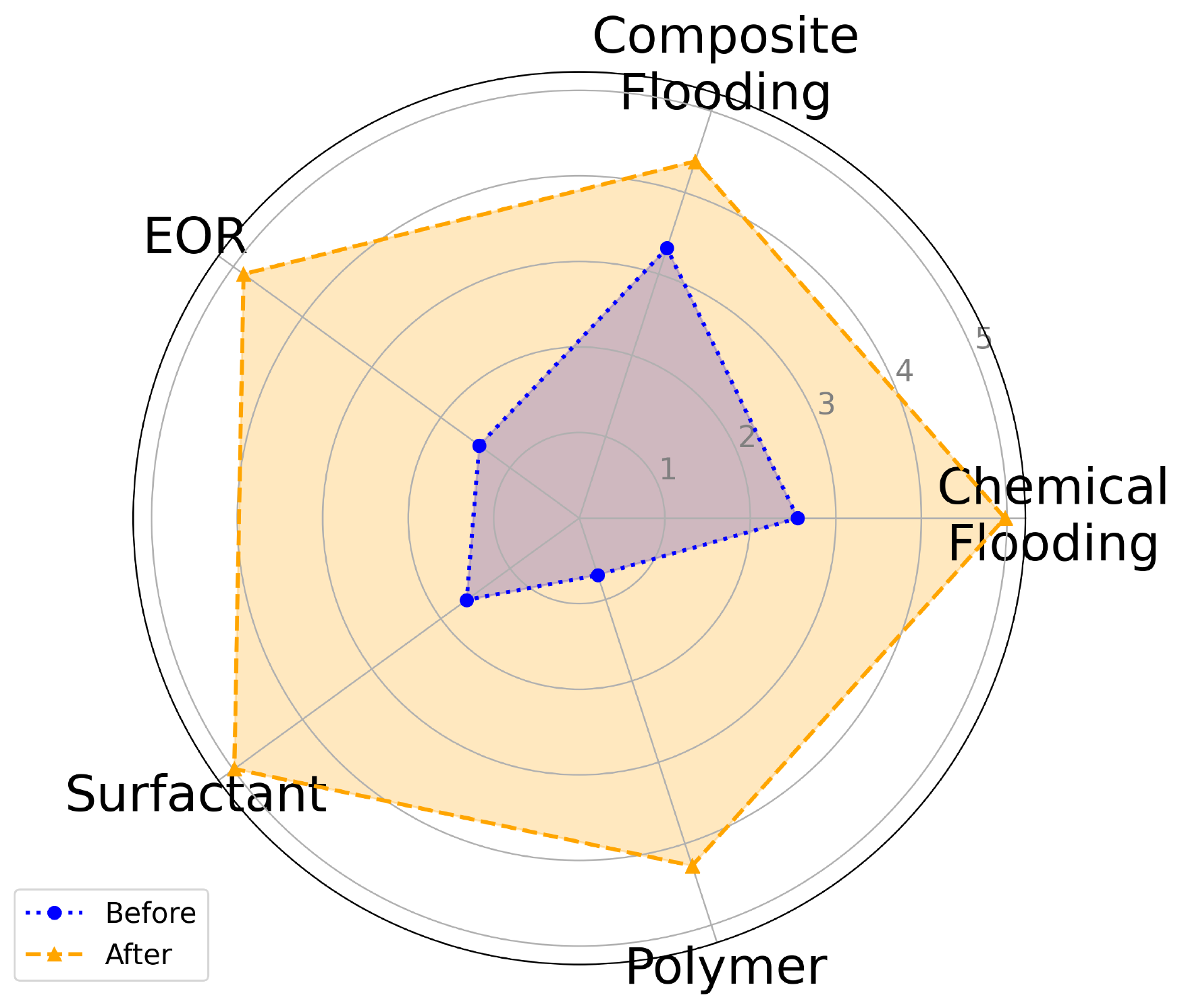}
    \includegraphics[width=0.3\linewidth]{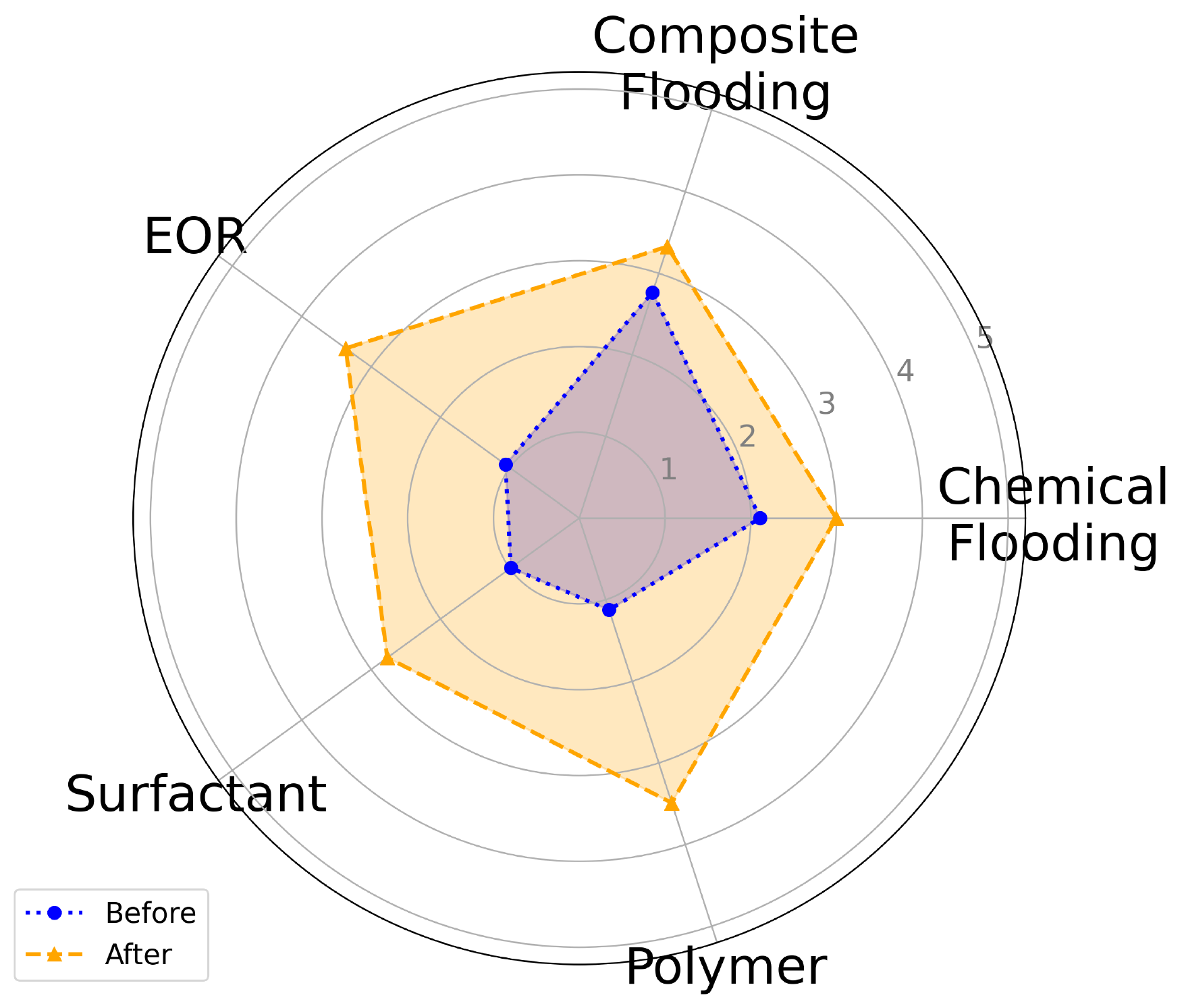}
    \includegraphics[width=0.3\linewidth]{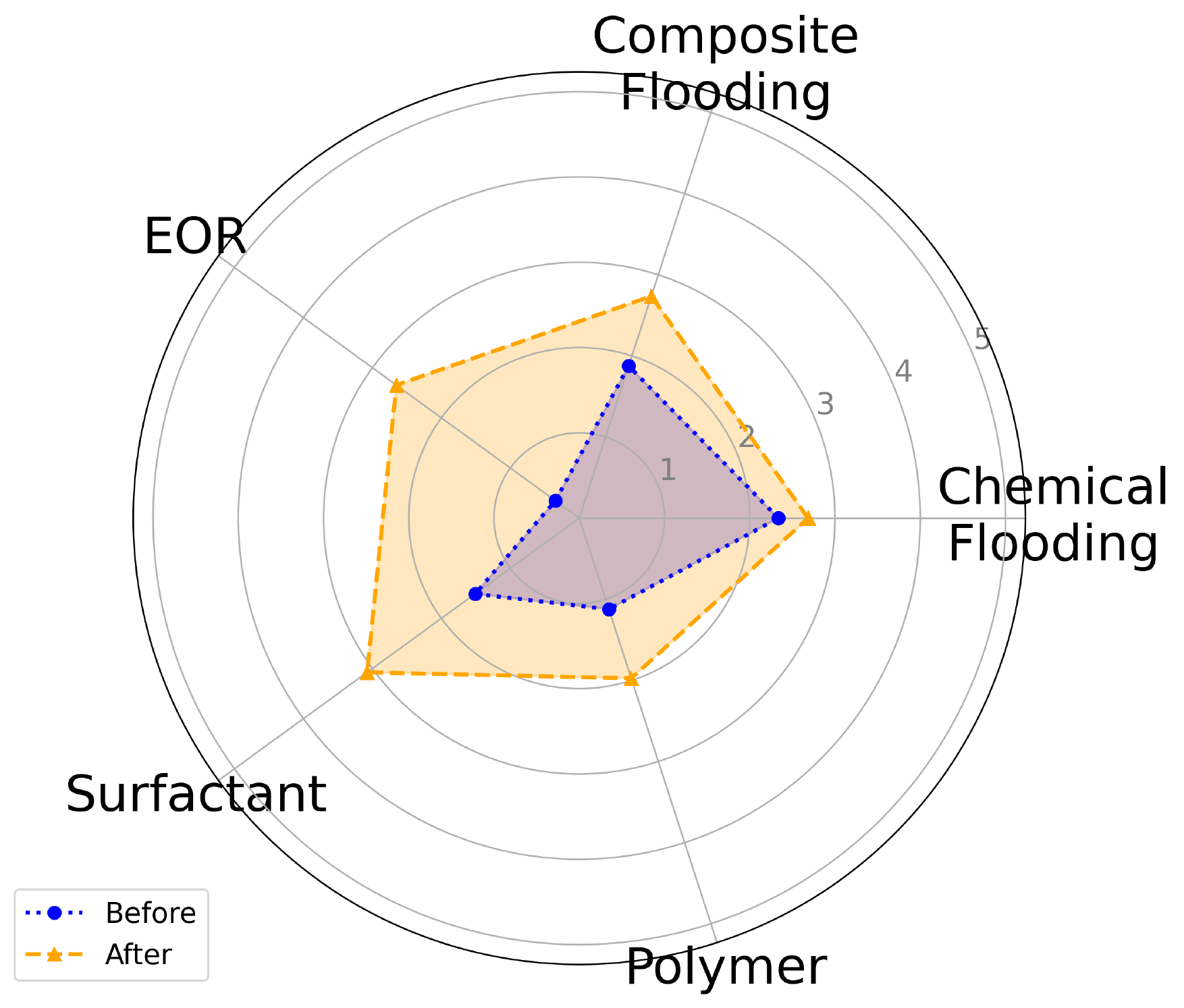}
    \caption{Evolution of user knowledge states ($\boldsymbol{\theta}$) across five key concepts using different guiding question generation methods. From left to right: AGQ, Chain-of-Thought (CoT), and Zero-shot. The results demonstrate the effectiveness of the AGQ framework's generated questions in enhancing user understanding.}
    \label{fig:all_thetas}
\end{figure*}

\section{Results and Analysis}

\subsection{Comparison with Baselines} \label{comparison-with-baselines}

In this section, we evaluate the performance of the AGQ framework against several baseline methods and assess its generalizability across different LLMs. The primary comparison involves AGQ, Zero-shot Question Generation, CoT Prompts with Handcrafted Examples, and a Human Expert benchmark. The accuracy is evaluated through a systematic process: we first collect sets of conversations generated by each method, then employ an LLM to answer questions from the EOR-QA dataset based on the conversation context. Accuracy is defined as the average rate of correct responses.

Figure \ref{fig:baselines' accuracy} illustrates the accuracy comparison over dialogue rounds. The results demonstrate AGQ's superior performance. After 20 rounds, AGQ achieved 100\% accuracy, significantly higher than CoT (41.1\%) and Zero-shot (23.9\%). Across all rounds, AGQ maintained a higher average accuracy (48.8\%) compared to CoT (25.6\%) and Zero-shot (16.3\%). The performance trajectory of AGQ closely mirrors that of the "Human with AGQ context" group (average 51.0\%, reaching $\ge$95\% accuracy by round 19), validating our automatic evaluation methodology, and approaches the effectiveness of human experts (average 72.0\%, reaching $\ge$95\% accuracy by round 14).

While CoT demonstrates strong logical reasoning, its accuracy improvement was limited by susceptibility to prompt examples, leading to repetitive questions and reduced information retrieval efficiency. The Zero-shot method, lacking context or examples, generated vague questions, resulting in consistently low accuracy with minimal variance. In contrast, AGQ generates more specific and complex questions compared to the baselines. See Table \ref{tab:qualitative-examples} for examples of guiding questions generated by each method addressing a similar underlying theme.

\begin{table}[ht]
    \centering
    \caption{Qualitative Comparison of Example Guiding Questions}
    \label{tab:qualitative-examples}
    \begin{tabular}{lp{0.8\linewidth}} \\ 
        \toprule
        Method & Guiding Question \\
        \midrule
        AGQ & How do surfactants and hydrocarbon miscible flooding techniques synergistically Enhance Oil Recovery during the process of increasing oil recovery? \\
        \midrule
        CoT & What is the principle of hydrocarbon miscible flooding? \\
        \midrule
        Zero-shot & How does the principle or technology mentioned in this conversation work? \\
        \bottomrule
    \end{tabular}
\end{table}

Furthermore, to assess the generalizability of the AGQ framework, its performance was evaluated using distinct LLMs: Qwen2.5-7B and Qwen2.5-32B \cite{qwen2025qwen25technicalreport}. Figure \ref{fig:diff_llm_accuracy} presents this cross-model comparison. The results illustrate several key characteristics. Firstly, initial performance correlates with model scale. In the early rounds (e.g., rounds 1-10), the larger Qwen2.5-32B model demonstrates higher accuracy compared to the smaller Qwen2.5-7B and ChatGLM4-9B models. For instance, at round 7, Qwen2.5-32B achieved approximately 50\% accuracy, whereas Qwen2.5-7B registered below 40\%. Secondly, despite these initial differences, accuracy consistently improves with increasing interaction rounds across all tested LLMs. Critically, all models converge to high accuracy levels (approaching 100\%) by rounds 19-21. This convergence indicates the framework's capacity to effectively enhance the performance of LLMs regardless of their initial capacity. Collectively, these findings confirm AGQ's superiority over baseline methods and highlight its cross-model adaptability and robustness. Its effectiveness is not contingent on a specific LLM, showcasing its potential as a generalizable approach for enhancing information retrieval guidance systems across various model choices.

\subsection{Text Similarity Evaluation}

To complement our evaluation, we also conducted a quantitative analysis comparing the guiding questions generated by each method against human expert reference questions using standard text similarity metrics \cite{papineni2002bleu,lin2004rouge}. While these metrics primarily assess lexical and semantic similarity rather than functional effectiveness in information retrieval, they provide an additional perspective on performance comparison. As shown in Table~\ref{tab:bleu-rouge-eval}, the AGQ framework substantially outperformed baseline methods across all metrics. AGQ achieved a BLEU-4 score of 0.219, significantly higher than CoT (0.025) and Zero-shot (0.016). Similarly, for ROUGE-1, AGQ scored 0.577 compared to 0.198 for CoT and 0.114 for Zero-shot. This superior performance extended to ROUGE-2 (AGQ: 0.278, CoT: 0.023, Zero-shot: 0.008) and ROUGE-L (AGQ: 0.463, CoT: 0.168, Zero-shot: 0.102). This result reflects the characteristic of question generation tasks, as functionally equivalent questions can be expressed using varied lexical choices and syntactic structures. Despite this inherent variability, AGQ\'s substantially higher scores confirm its greater alignment with human expert formulations. Full details of this evaluation can be found in Appendix.

\begin{table}[ht]
    \centering
    \caption{Quantitative Evaluation using Text Similarity Metrics against Human Expert References}
    \label{tab:bleu-rouge-eval}
    \begin{tabular}{lcccc}
        \toprule
        Method & BLEU-4 & ROUGE-1 & ROUGE-2 & ROUGE-L (F1) \\
        \midrule
        AGQ & 0.219 & 0.577 & 0.278 & 0.463 \\
        CoT & 0.025 & 0.198 & 0.023 & 0.168 \\
        Zero-shot & 0.016 & 0.114 & 0.008 & 0.102 \\
        \bottomrule
    \end{tabular}
\end{table}

\begin{figure}[t]
    \centering
    \includegraphics[width=1\columnwidth]{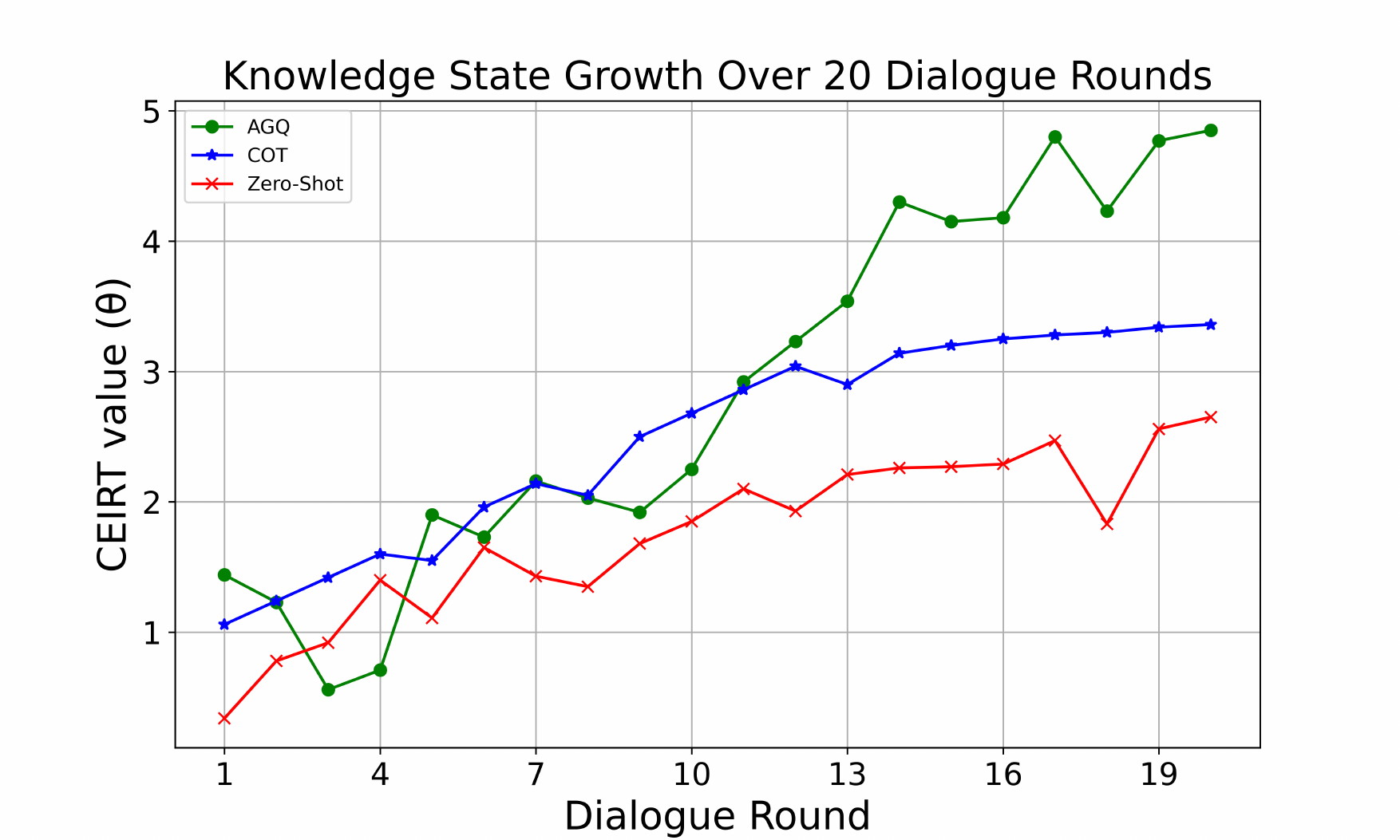}
    \caption{After 20 rounds of dialogue, the knowledge state ($\theta_j$) value of EOR increased from 1.44 to 4.85 with AGQ, from 1.06 to 3.36 with CoT, and from 0.34 to 2.65 with Zero-shot.}
    \label{fig:theta_per_rounds}
\end{figure}

\subsection{Knowledge Gain}
To evaluate the impact on user understanding, we evaluated knowledge gain by comparing the knowledge state vector ($\boldsymbol{\theta}$) before and after the 20-round interaction process. 

Figure \ref{fig:theta_per_rounds} tracks the knowledge state ($\theta_j$) for the EOR concept specifically. It shows that the AGQ framework facilitated substantial and steady growth, increasing the $\theta_j$ value from 1.44 to 4.85. This contrasts sharply with the moderate increase observed for CoT (1.06 to 3.36) and the minimal gain for the Zero-shot method (0.34 to 2.65), underscoring AGQ's effectiveness in deepening understanding of specific concepts.

Further evidence is presented in Figure \ref{fig:all_thetas}, which displays the final knowledge states across five concepts in EOR-QA using radar charts. Visually, AGQ demonstrates superior performance across all concepts compared to CoT and Zero-shot methods. Quantitatively, the average final knowledge state across these five concepts was 4.70 for AGQ, significantly higher than 3.19 for CoT and 2.63 for Zero-shot. This substantial knowledge gain observed with AGQ aligns with the superior accuracy reported in Section \ref{comparison-with-baselines}, suggesting that the framework's ability to generate effective guiding questions translates directly into improved user understanding, retrieval efficiency, and perceived guidance quality.

\subsection{Analysis}
In comparison, while the CoT method demonstrates capability in guiding users through the learning process, it lacks the precision needed to effectively address specific concepts where the user's understanding is weaker. This limitation stems from its inability to differentiate adequately between the user's varying levels of knowledge state, resulting in a less targeted approach to information retrieval. On the other hand, the Zero-shot question generation method shows even weaker performance. Without accurate knowledge state measurement, the Zero-shot method generates overly general guiding questions lacking specific content.

The superior performance of the AGQ framework can be attributed to the integration of the CEIRT model, which plays a crucial role in its success. The CEIRT model enables dynamic and precise knowledge measurement. This model-driven precision enables AGQ to generate guiding questions that are not only relevant but also strategically designed to challenge and expand the user's knowledge in targeted concepts. By dynamically updating the user's knowledge state and refining the difficulty and discrimination parameters (i.e., $\boldsymbol{\theta}$, $\boldsymbol{b}$, and $\boldsymbol{a}$), AGQ improves the effectiveness of information retrieval.

\section{Ablation Study} \label{ablation}

\begin{figure}[ht]
    \centering
    \includegraphics[width=1\columnwidth]{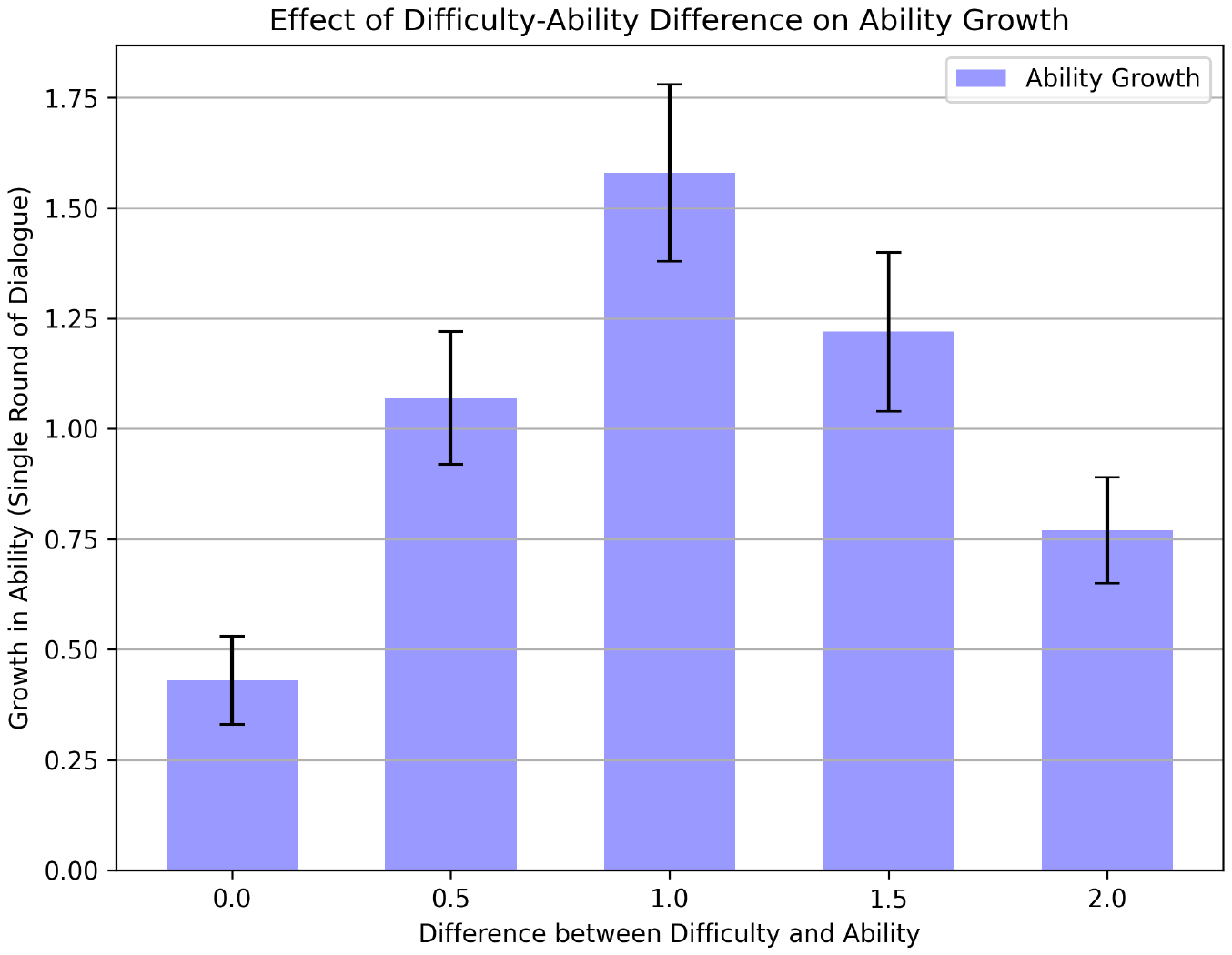}
    \caption{Ability growth peaks when the difficulty-knowledge gap ($b_i - \theta_j$) is approximately 1, suggesting optimal learning occurs at a moderate challenge. Error bars denote variability.}
    \label{fig:ablation}
\end{figure}

This section investigates the optimal difference between the difficulty of question $i$ ($b_i$) and the user's knowledge state in concept $j$ ($\theta_j$) by experimentally comparing knowledge state growth under varying difficulty-ability differences.
As illustrated in Figure \ref{fig:ablation}, knowledge gain is maximized when the absolute difference between question difficulty $b_i$ and user knowledge state $\theta_j$ is approximately 1. This finding provides empirical support for the design of our Inspiring Text suitability score (Equation \ref{eq:suitability_score}) and underlines the importance of providing an appropriate level of challenge to optimize information retrieval outcomes. Deviations from this difference significantly reduced knowledge gain, emphasizing the need to align question difficulty with user knowledge.
Furthermore, as the difference was adjusted to 1.5, users began to report that the difficulty gap was significant, with a noticeable decline in engagement. This feedback became more pronounced as the difference increased to 2.

\section{Conclusion}

In this paper, we introduced the AGQ framework, an innovative approach for generating guiding questions during the information retrieval process. At its core, the framework leverages the CEIRT model, a novel method for measuring knowledge in specialized domains that enables precise assessment of users' knowledge states across multiple concepts.
In comparative experiments against several baseline methods, AGQ demonstrated superior performance, closely matching the effectiveness of human experts. 
Additionally, we developed a domain-specific text dataset (EOR-QA), which served as both a source of structured knowledge and inspiring text to facilitate precise guiding question generation within the framework during our evaluations. 
The AGQ framework marks a key advance in LLM-based adaptive guidance, offering a versatile design that is readily adaptable to knowledge-intensive domains like medicine and education beyond its initial demonstration in the EOR field.

\bibliography{references}

\section*{Acknowledgements}
This work is supported by the Science Foundation of China University of Petroleum, Beijing (Grant No. 2462023YJRC024) and the Frontier Interdisciplinary Exploration Research Program of China University of Petroleum, Beijing (Grant No. 2462024XKQY003). Zhongqi Lu is the corresponding author.

\clearpage
\section{Appendix}
\input{Supplementary-Material}

\end{document}

%% file: Supplementary-Material.tex
\appendix
\setcounter{table}{0}
\setcounter{figure}{0}
\setcounter{equation}{0}
\renewcommand{\thetable}{A\arabic{table}}
\renewcommand{\thefigure}{A\arabic{figure}}
\renewcommand{\theequation}{A\arabic{equation}}

\begin{figure*}[ht]
    \centering
    \includegraphics[width=0.9\linewidth]{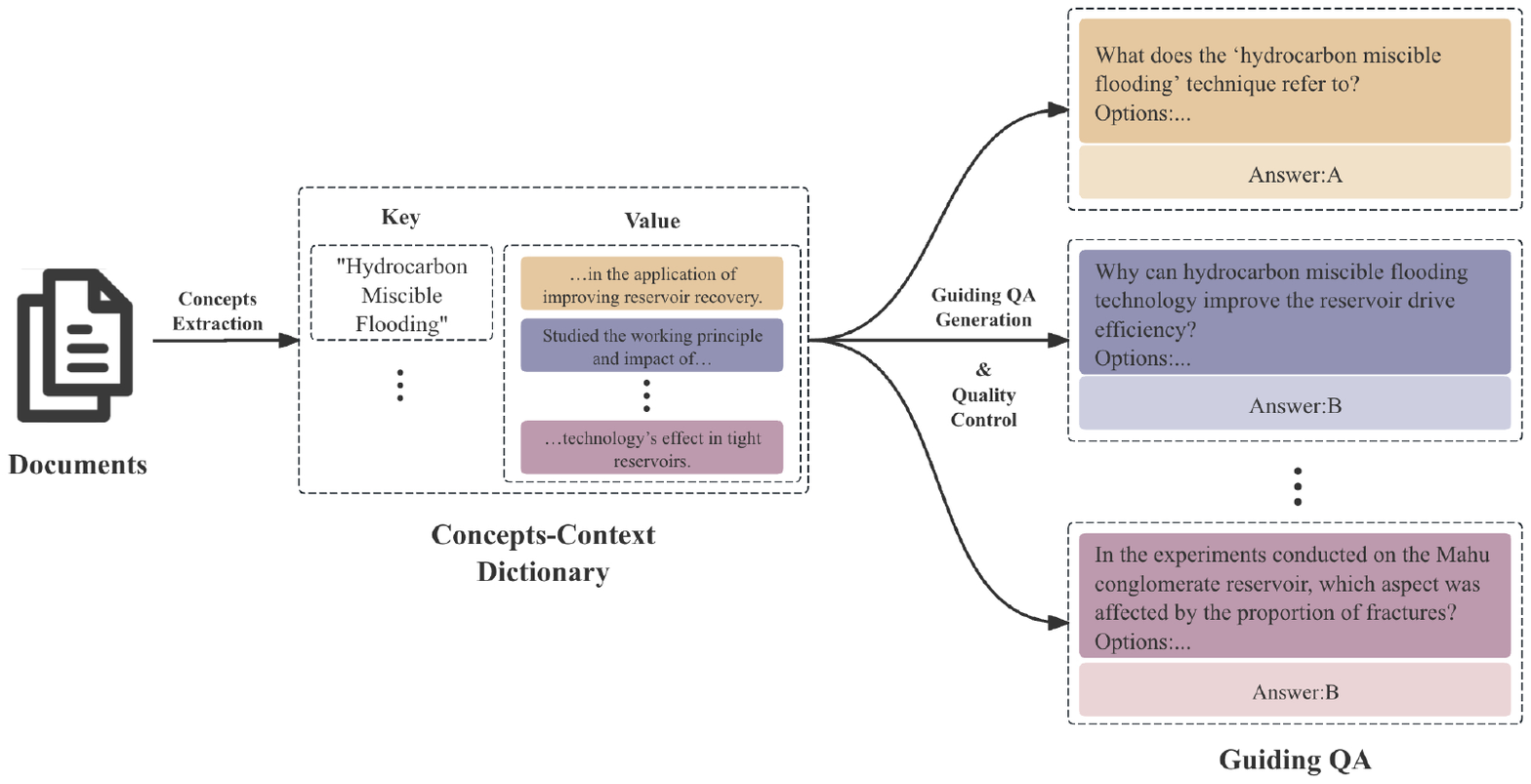}
    \caption{ This diagram illustrates the EOR-QA generation process using hydrocarbon miscible flooding as an example, including the extraction of concepts, the construction of a Concepts-Context Dictionary, and the subsequent generation and quality control of guiding questions.}
    \label{fig:data-set-generation}
\end{figure*}

\begin{table*}[!ht]
    \centering
    \caption{Examples of Dataset}
    \begin{tabularx}{\textwidth}{X|X|X|X|X}
        \toprule
        \textbf{Content} & \textbf{Options} & \textbf{Answer} & \textbf{Concept} & \textbf{Scenarios}\\
        \midrule 
        In chemical flooding, which metric significantly changes in third type vs. second type? & 
        a) Surfactant adsorption loss\newline
        b) Alkali adsorption\newline
        c) Polymer adsorption\newline
        d) Oil displacement efficiency & 
        a) &
        Third type oil reservoirs &
        Application\\
        \midrule
        In petroleum geology, what are 'Third type oil reservoirs'? &
        a) Low permeability\newline
        b) Poor properties and low pressure\newline
        c) High permeability with good properties\newline
        d) High crude oil sulfur content &
        b) &
        Third type oil reservoirs &
        Theory\\
        \midrule
        In Daqing Oilfield tests, how many percentage points higher is surfactant loss in third vs. second type? &
        a) 32.1\%\newline
        b) 52.1\%\newline
        c) 37.5\%\newline
        d) 42.6\% & b) &
        Third type oil reservoirs &
        / \\
        \bottomrule
    \end{tabularx}
    \label{tab:examples-of-dataset}
\end{table*}

\begin{table*}[ht]
    \centering
    \caption{Dataset Comparison}
    \begin{tabular}{l|c|c|c|c}
        \toprule
        \textbf{Datasets} & \textbf{Size} & \textbf{Max length} & \textbf{Granularity} & \textbf{Scenarios} \\
        \midrule
        EduQG & 3400 & 227 & \ding{55} & \ding{55} \\
        \midrule
        MoocRadar(Math) & 227 & 100 & 50.12 & \ding{51} \\
        \midrule
        SciQ & 11680 & 77 & \ding{55} & \ding{55} \\
        \midrule
        SXT001\_CN & 23990 & 274 & \ding{55} & \ding{55} \\
        \midrule
        ai2\_arc & 10309 & 683 & \ding{55} & \ding{55} \\
        \midrule
        EOR-QA & 3142 & 300 & 120.84 & \ding{51} \\
        \bottomrule
    \end{tabular}
    \label{tab:dataset_info}
\end{table*}

\section{EOR-QA Dataset}
The EOR-QA (Enhanced Oil Recovery Question-Answer) dataset was developed to provide the structured knowledge required by the AGQ framework, addressing the knowledge limitations of general large language models in the specialized domain of Enhanced Oil Recovery. As shown in Figure \ref{fig:data-set-generation}, this dataset was constructed through a multi-stage pipeline involving LLM-based concept identification, sentence extraction, QA pair generation, followed by manual verification by domain experts.

\subsection{Generation Process}
During the concept extraction phase, key concepts were identified from documents related to the EOR topics. To ensure accuracy and relevance, the concepts were validated manually. After validation, these concepts were linked with their contextual information to create a concept-context dictionary, providing a solid foundation for generating multiple-choice questions. This dictionary ensured that the questions were grounded in comprehensive background information, maintaining their scientific rigor. With this dictionary, over 3,100 multiple-choice questions were generated, each with four options and a correct answer, ensuring that the questions remained closely aligned with the original text, thus preserving their authority and quality.

\subsection{Comparison}
As shown in Table \ref{tab:dataset_info}, we compared several datasets with EOR-QA: 
\begin{itemize}
	\item \textbf{EduQA:} EduQA is a educational dataset with 3,397 samples, including multiple-choice questions, answers, and distractors linked to source documents. It is enhanced with Bloom's taxonomy cognitive complexity labels for 903 questions, all crafted by experts. 
	\item \textbf{MoocRadar-Math:} MoocRadar is an extensive knowledge repository with 2,513 exercise questions, offering detailed annotations for student learning characteristics and cognitive states. For comparative experiments, we have extracted the mathematics-related portion of this dataset.
	\item \textbf{SciQ:} SciQ dataset contains 13,679 crowdsourced science multi-choice questions about Physics, Chemistry and Biology, among others. For the majority of the questions, an additional paragraph with supporting evidence for the correct answer is provided.
    \item \textbf{STX001\_cn:} This extensive math exam dataset with multi-choice questions from primary to high school, is ideal for educational research and AI system development, available on Alibaba Cloud.
	\item \textbf{ai2\_arc:} The ai2\_arc dataset, designed for advanced question-answering research in grade-school science, offers 7,787 multi-choice questions across Challenge and Easy Sets.
\end{itemize}
Overall, EOR-QA, as a domain-specific dataset, achieves the highest coverage of useful information. When considering the number of questions per concept as a measure of granularity, EOR-QA demonstrates a significant advantage over similar datasets. On average, each concept in EOR-QA contains 120.84 questions. We also classified the questions into two levels of difficulty—understanding and application—based on the CEIRT model training.
To our knowledge, there are currently very few question-answering datasets in the petroleum field. 
The introduction of EOR-QA not only contributes to this critical gap but also facilitates the development of specific-domain LLMs that are better suited to the petroleum industry. While further work is needed, this dataset offers a valuable starting point for creating models that can more effectively address the specific challenges of this domain.

\section{Details of Prompt Engineering}
In the following subsections, prompts will be presented in \textit{italics}.
\subsection{Examples in EOR-QA Filtering Process}

The following examples are used for in-context training to help the LLM distinguish whether a question involves specific experiments:

\textbf{Question:} In chemical flooding technology, which additive can significantly improve the displacement efficiency of the formation?
    
\textbf{Experiment-related:} \textbf{no}
    
\textbf{Question:} In the chemical flooding technology of oil fields, which method does not achieve improved displacement efficiency by adjusting the capillary number?
    
\textbf{Experiment-related:} \textbf{no}
    
\textbf{Question:} In chemical flooding in oil fields, which driving method has sufficient formulation flexibility to cope with the diversity of geology and reservoirs?
    
\textbf{Experiment-related:} \textbf{no}
    
\textbf{Question:} How does Daqing Oilfield distinguish and handle different types of oil layers when implementing weak alkali ternary composite flooding?
    
\textbf{Experiment-related:} \textbf{yes}
    
\textbf{Question:} In the Upper Wuerhe Formation glutenite reservoir of the Mahu Well Area, what impact might 'water sensitivity damage' have on the exploitation of the reservoir?
    
\textbf{Experiment-related:} \textbf{yes}
    
\textbf{Question:} In the on-site chemical flooding test of Daqing Oilfield, how many percentage points higher is the surfactant adsorption loss in type III oil layers compared to type II oil layers?
    
\textbf{Experiment-related:} \textbf{yes}

\subsection{Prompt in User-LLM Interaction}
The following is the prompt used during the user-LLM interaction process:

\textit{You are deeply involved in the field of enhancing oil recovery (EOR). Your task is to teach users based on their individual knowledge state for each concept:\textbf{\{Knowledge State\}}, helping them learn the material and apply it in practical scenarios. When responding to users' questions, provide guidance that directly addresses the question asked, without offering additional information. Keep your answers as concise as possible. There's no need to recommend learning steps or resources, just impart the knowledge.}

\subsection{Prompt in Guiding Question Generation}

\subsubsection{Low Knowledge State Prompt ($P_{QGlow}$)}

\textit{You are an expert in the field of Enhanced Oil Recovery (EOR). You are now required to propose 5 guiding questions based on the following knowledge points: \textbf{\{conversation\_concepts\}}. Your primary goal is to help the user clarify fundamental principles, mechanisms, and definitions related to these concepts, based on the following text: \textbf{\{Inspiring\_Text\}}. Ensure the questions target foundational understanding. Additionally, you can refer to the following example:}

\begin{itemize}
    \item \textit{``How do surfactants and hydrocarbon miscible flooding techniques synergistically Enhance Oil Recovery during the process of increasing oil recovery?''}

    \item \textit{``In enhanced oil recovery technologies, how are steam injection and hydrocarbon miscible flooding combined to optimize the displacement efficiency of reservoirs?''}

    \item \textit{``Discuss the application scenarios of CO2 in enhanced oil recovery, particularly its synergistic effect when combined with polymer flooding, considering the physical and chemical properties of CO2.''}

    \item \textit{``How do the physical and chemical properties of CO2 affect its effectiveness as a displacement agent in CO2 flooding, especially in the synergistic mechanism when combined with polymer flooding?''}

    \item \textit{``How does nuclear magnetic resonance (NMR) technology reflect changes in fluid distribution within the reservoir when evaluating recovery efficiency? Analyze how the physical and chemical properties of steam injection and surfactants impact recovery efficiency.''}

    \item \textit{``How does multi-field reconstruction displacement technology utilize various methods to synergistically improve the development effectiveness of tight reservoirs? Specifically, what technical measures are involved, and what are their mechanisms of action?''}

    \item \textit{``How can the degree of oil recovery be improved?''}

    \item \textit{``Based on the characteristics of different types of reservoirs, how can suitable enhanced oil recovery techniques (such as hydrocarbon miscible flooding, steam injection, etc.) be selected to improve recovery efficiency, and what are the application scenarios of these technologies?''}

    \item \textit{``In enhanced oil recovery technologies, how can suitable displacement techniques (such as hydrocarbon miscible flooding, steam injection, etc.) be selected based on the characteristics of different types of reservoirs, and what are the application scenarios and physical and chemical properties of these technologies?''}

    \item \textit{``For different types of reservoirs, how can suitable enhanced oil recovery techniques (such as hydrocarbon miscible flooding, steam injection, etc.) be selected to improve recovery efficiency? Please analyze the application scenarios and physical and chemical properties of these technologies.''}

    \item \textit{``How can thermal composite development technologies be reasonably selected and applied in enhanced oil recovery, and how can their effectiveness be evaluated when combined with other methods like steam injection, particularly in terms of applicability and physical and chemical properties for different reservoir types?''}

    \item \textit{``For different types of reservoirs (such as Class III reservoirs), how can suitable enhanced oil recovery techniques (such as hydrocarbon miscible flooding, steam injection, etc.) be selected based on their characteristics to improve recovery efficiency? Please analyze the application scenarios and physical and chemical properties of these technologies.''}
\end{itemize}

\textit{Do not provide any output other than the guiding questions, and ensure the given knowledge points are included in the guiding questions.}

\subsubsection{High Knowledge State Prompt ($P_{QGhigh}$)}

\textit{You are an expert in the field of Enhanced Oil Recovery (EOR). You are now required to propose 5 guiding questions based on the following knowledge points: \textbf{\{conversation\_concepts\}}. Your primary goal is to guide the user to explore practical applications, compare scenarios, or synthesize information related to these concepts, based on the following text: \textbf{\{Inspiring\_Text\}}. Ensure the questions target advanced application or deeper integration of knowledge. Additionally, you can refer to the following example:}

\begin{itemize}
    \item \textit{``How do surfactants and hydrocarbon miscible flooding techniques synergistically Enhance Oil Recovery during the process of increasing oil recovery?''}

    \item \textit{``In enhanced oil recovery technologies, how are steam injection and hydrocarbon miscible flooding combined to optimize the displacement efficiency of reservoirs?''}

    \item \textit{``Discuss the application scenarios of CO2 in enhanced oil recovery, particularly its synergistic effect when combined with polymer flooding, considering the physical and chemical properties of CO2.''}

    \item \textit{``How do the physical and chemical properties of CO2 affect its effectiveness as a displacement agent in CO2 flooding, especially in the synergistic mechanism when combined with polymer flooding?''}

    \item \textit{``How does nuclear magnetic resonance (NMR) technology reflect changes in fluid distribution within the reservoir when evaluating recovery efficiency? Analyze how the physical and chemical properties of steam injection and surfactants impact recovery efficiency.''}

    \item \textit{``How does multi-field reconstruction displacement technology utilize various methods to synergistically improve the development effectiveness of tight reservoirs? Specifically, what technical measures are involved, and what are their mechanisms of action?''}

    \item \textit{``How can the degree of oil recovery be improved?''}

    \item \textit{``Based on the characteristics of different types of reservoirs, how can suitable enhanced oil recovery techniques (such as hydrocarbon miscible flooding, steam injection, etc.) be selected to improve recovery efficiency, and what are the application scenarios of these technologies?''}

    \item \textit{``In enhanced oil recovery technologies, how can suitable displacement techniques (such as hydrocarbon miscible flooding, steam injection, etc.) be selected based on the characteristics of different types of reservoirs, and what are the application scenarios and physical and chemical properties of these technologies?''}

    \item \textit{``For different types of reservoirs, how can suitable enhanced oil recovery techniques (such as hydrocarbon miscible flooding, steam injection, etc.) be selected to improve recovery efficiency? Please analyze the application scenarios and physical and chemical properties of these technologies.''}

    \item \textit{``How can thermal composite development technologies be reasonably selected and applied in enhanced oil recovery, and how can their effectiveness be evaluated when combined with other methods like steam injection, particularly in terms of applicability and physical and chemical properties for different reservoir types?''}

    \item \textit{``For different types of reservoirs (such as Class III reservoirs), how can suitable enhanced oil recovery techniques (such as hydrocarbon miscible flooding, steam injection, etc.) be selected based on their characteristics to improve recovery efficiency? Please analyze the application scenarios and physical and chemical properties of these technologies.''}
\end{itemize}

\textit{Do not provide any output other than the guiding questions, and ensure the given knowledge points are included in the guiding questions.}

\subsubsection{CoT Prompt}
In the CoT method, we use the following prompt:

\textit{Following the provided examples, generate five guiding questions:}
\begin{itemize}
    \item ``\textit{\textbf{Query:}I want to learn about chemical EOR and related knowledge'',
    ``\textbf{Response:}Chemical Enhanced Oil Recovery (EOR) involves injecting chemicals to alter reservoir fluid properties and Enhance Oil Recovery. Key types include alkaline flooding, polymer flooding, and surfactant flooding. Study related fields like reservoir engineering, chemical engineering, and geology. Suggested resources include textbooks, case studies, lab experiments, and online courses.'',
    ``\textbf{Reasoning:} Given the interest in chemical EOR, the following guiding questions could explore related EOR techniques or expand on specific aspects of chemical EOR.'',
    ``\textbf{Guiding Questions:}1. What is hydrocarbon miscible flooding?2. What are the principles of thermal recovery?3. What is the role of steam injection in oilfield development?4. What are the steps in thermal composite development?5. What are the characteristics and effects of weak alkali composite flooding?''}

    \item \textit{``\textbf{Query:} What is the role of steam injection in oilfield development?'',
    ``\textbf{Response:} Steam injection enhances oil recovery by heating the crude, reducing viscosity, and improving fluid flow. It also alters rock permeability and provides additional drive pressure.'',
    ``\textbf{Reasoning:} The following questions could focus on other thermal recovery methods or specific effects of steam injection.'',
    ``\textbf{Guiding Questions:} 1. What is hydrocarbon miscible flooding?2. What are the main thermal recovery methods?3. How does steam injection improve oil recovery?4. What are the steps in thermal composite development?5. What are the characteristics of weak alkali composite flooding?''}

    \item \textit{``\textbf{Query:}What are the steps in thermal composite development?'',
    ``\textbf{Response:}Thermal composite development combines thermal recovery with chemical EOR. Steps include reservoir evaluation, design, steam injection, chemical injection, and continuous monitoring.'',
    ``\textbf{Reasoning:} Reasoning: The following questions could explore specific techniques within thermal composite development or compare it to other EOR methods.'',
    ``\textbf{Guiding Questions:} 1. What is hydrocarbon miscible flooding?2. What principles support thermal recovery?3. How is steam injection applied in oil extraction?4. What are the advantages of thermal composite development?5. What are the characteristics of weak alkali composite flooding?''}
\end{itemize}

\subsubsection{Zero-shot Prompt}
In the Zero-shot method, we use the following prompt:

\textit{Based on a given human-machine dialogue, propose five guiding questions focused on the user's weak areas: \textbf{\{knowledge state\}}, aiming to guide the LLM to answer about principles or applications. You are not allowed to output anything else.}

\section{Sample Guiding Questions}
Below are examples of guiding questions generated during the experiments by our AGQ framework:
\begin{itemize}
    \item How can suitable EOR techniques be selected based on reservoir characteristics (e.g., fractures, surfactants), and what are the specific application scenarios and physical-chemical properties of these techniques in enhancing recovery rates?
    \item Why does the extraction rate of shale oil in fractures and surrounding areas initially increase and then slow down with extended CO2 injection? Analyze this trend considering the physical properties of fractures and the chemical characteristics of CO2.
    \item How can the properties of non-condensable gases (NCG) at high temperatures be utilized, along with the use of fractures and surfactants, to enhance the recovery of heavy oil reservoirs during enhanced oil recovery?
    \item How can NMR technology assess changes in recovery rates during EOR? Specifically, how do steam injection and surfactants affect fluid distribution and recovery rates in the reservoir?
    \item What is the mechanism of action of surfactants in enhanced oil recovery, and how do they synergize with hydrocarbon miscible flooding to reduce crude oil viscosity and improve recovery efficiency?
    \item How can the physical and chemical properties of CO2, particularly when combined with polymer flooding, be utilized to analyze its solubility, diffusion, and solvation capabilities and their impact on recovery efficiency in enhanced oil recovery?
\end{itemize}

\section{Human Evaluation}

To ensure the validity of our results, we conducted a human evaluation of the guiding questions generated by our proposed method (AGQ) and two baseline methods. We recruited 10 independent reviewers with varying levels of expertise in the petroleum field to assess the generated questions across different knowledge backgrounds. Each method was evaluated on three key dimensions:
\begin{itemize}
    \item \textbf{Diversity:} The richness and non-repetitiveness of the questions.
    \item \textbf{Relevance:} The alignment of the questions with the user's knowledge gap and target concepts.
    \item \textbf{Guidance:} The effectiveness of the questions in facilitating information retrieval and learning.
\end{itemize}

The aggregated evaluation scores are presented in Table~\ref{tab:human-verify}, demonstrating the superior performance of our AGQ framework.

\begin{table}[h!]
    \centering
    \caption{Human evaluation scores for the three question generation methods. AGQ significantly outperforms the baselines across all dimensions.}
    \label{tab:human-verify}
    \begin{tabular}{cccc}
        \toprule
        Method & Diversity & Relevance & Guidance \\
        \midrule
        Zero-shot & 12 & 0 & 56 \\
        CoT & 64 & -59 & 72 \\
        AGQ & 87 & 87 & 95 \\
        \bottomrule
    \end{tabular}
\end{table}

Below are qualitative examples of the guiding questions generated by each method. 
\subsection{AGQ}
\begin{itemize}
    \item How do surfactants and hydrocarbon miscible flooding techniques synergistically Enhance Oil Recovery during the process of increasing oil recovery?

    \textbf{Diversity:}1 \textbf{Relevance:}1 \textbf{Guidance:}1

    \item In enhanced oil recovery technologies, how are steam injection and hydrocarbon miscible flooding combined to optimize the displacement efficiency of reservoirs?

    \textbf{Diversity:}1 \textbf{Relevance:}1 \textbf{Guidance:}1

    \item How does the use of surfactants reduce interfacial tension in the process of enhancing oil recovery, and how does this effect influence the recovery efficiency in fractured reservoirs?

    \textbf{Diversity:}1 \textbf{Relevance:}1 \textbf{Guidance:}1

    \item How can the comprehensive effects of different displacement fluids (such as CO2, polymers, etc.) on enhancing the recovery rate of the Mahu tight conglomerate reservoir be evaluated through optimized orthogonal experimental design during the implementation of fracturing network technology?

    \textbf{Diversity:}1 \textbf{Relevance:}1 \textbf{Guidance:}1

    \item How can suitable enhanced oil recovery techniques be selected based on the characteristics of different types of reservoirs (such as fractures, surfactants, etc.)?

    \textbf{Diversity:}0 \textbf{Relevance:}1 \textbf{Guidance:}1

    \item How can the effectiveness of hydrocarbon miscible flooding be optimized by adjusting the fracture structure during the process of enhancing oil recovery?

    \textbf{Diversity:}1 \textbf{Relevance:}1 \textbf{Guidance:}1

    \item How can thermal composite development technology enhance the recovery rate of heavy oil reservoirs by altering reservoir temperature, physical properties, or driving force?

    \textbf{Diversity:}1 \textbf{Relevance:}1 \textbf{Guidance:}1

    \item How can the synergistic effect of polymer flooding systems and steam injection technology be utilized to enhance the recovery rate of the Daqing Oilfield in enhanced oil recovery techniques?

    \textbf{Diversity:}1 \textbf{Relevance:}0 \textbf{Guidance:}0

    \item Discuss the application mechanism of hydrocarbon miscible flooding in enhancing oil recovery in conjunction with steam injection technology.

    \textbf{Diversity:}1 \textbf{Relevance:}1 \textbf{Guidance:}1

    \item In CO2 flooding technology, how does temperature affect the solubility and diffusion capability of CO2, and how does this, in turn, influence its mechanism for enhancing oil recovery?

    \textbf{Diversity:}1 \textbf{Relevance:}1 \textbf{Guidance:}1
\end{itemize}
\subsection{CoT}
\begin{itemize}
    \item What are the principles and application scenarios of microbial flooding?

    \textbf{Diversity:}1 \textbf{Relevance:}-1 \textbf{Guidance:}1

    \item What is the working principle of electrochemical flooding, and how effective is its application in oilfield development?
    
    \textbf{Diversity:}1 \textbf{Relevance:}-1 \textbf{Guidance:}0

    \item What is the role of foam flooding technology in enhancing oil recovery?

    \textbf{Diversity:}1 \textbf{Relevance:}1 \textbf{Guidance:}1

    \item What are the applications of electrochemical flooding in oilfield development?

    \textbf{Diversity:}0 \textbf{Relevance:}-1 \textbf{Guidance:}1

    \item What is the principle of hydrocarbon miscible flooding?

    \textbf{Diversity:}0 \textbf{Relevance:}-1 \textbf{Guidance:}1

    \item What is the difference between cyclic steam stimulation and steam injection?

    \textbf{Diversity:}0 \textbf{Relevance:}-1 \textbf{Guidance:}1

    \item How can the economic benefits of thermal composite development be evaluated?

    \textbf{Diversity:}1 \textbf{Relevance:}-1 \textbf{Guidance:}0

    \item What challenges might be encountered during the implementation of thermal composite development, and what are the possible solutions?

    \textbf{Diversity:}1 \textbf{Relevance:}1 \textbf{Guidance:}1
    
    \item What is the principle of CO2 flooding technology, and how does it impact the environment?

    \textbf{Diversity:}1 \textbf{Relevance:}1 \textbf{Guidance:}0


\end{itemize}
\subsection{Zero-shot}
\begin{itemize}
    \item In which fields is this principle applied?

    \textbf{Diversity:}1 \textbf{Relevance:}0 \textbf{Guidance:}0 

    \item Can you provide examples of how this principle is applied in real-life situations?

    \textbf{Diversity:}1 \textbf{Relevance:}0 \textbf{Guidance:}0

    \item How does the principle or technology mentioned in this conversation work?

    \textbf{Diversity:}0 \textbf{Relevance:}0 \textbf{Guidance:}1

    \item In which fields has this principle been widely applied?

    \textbf{Diversity:}1 \textbf{Relevance:}0 \textbf{Guidance:}0

    \item What are the similarities and differences between this principle and other related principles or technologies?

    \textbf{Diversity:}0 \textbf{Relevance:}0 \textbf{Guidance:}1    

    \item What are the potential limitations or challenges of this principle?

    \textbf{Diversity:}0 \textbf{Relevance:}-1 \textbf{Guidance:}0    
\end{itemize}

\section{Human Expert Generated Questions}
\begin{itemize}
    \item Explain how the synergistic effects between surfactants and hydrocarbon miscible flooding techniques contribute to enhanced oil recovery processes.
    \item What mechanisms allow surfactants and hydrocarbon miscible flooding to work synergistically in the enhancement of oil recovery rates?
    
    \item Describe the methodologies for combining steam injection with hydrocarbon miscible flooding to optimize reservoir displacement efficiency in EOR applications.
    
    \item How does the integration of steam injection and hydrocarbon miscible flooding techniques optimize displacement efficiency in enhanced oil recovery operations?
    
    \item Analyze the mechanism by which surfactants reduce interfacial tension during oil recovery, with specific focus on how this affects recovery efficiency in fractured reservoir systems.
    
    \item What is the relationship between surfactant-induced interfacial tension reduction and recovery efficiency improvements in fractured reservoirs during EOR processes?
    
    \item Discuss methodologies for evaluating the comprehensive effects of displacement fluids like CO2 and polymers on recovery rates in tight conglomerate reservoirs using orthogonal experimental designs.
    
    \item How can orthogonal experimental design be utilized to assess the impact of various displacement fluids on enhancing recovery rates in tight conglomerate reservoirs?
    
    \item What criteria should guide the selection of appropriate enhanced oil recovery techniques based on specific reservoir characteristics including fracture networks and fluid properties?
    
    \item Outline the decision-making framework for selecting suitable EOR techniques based on different reservoir characteristics such as fracture patterns and wettability.
    
\end{itemize}

\section{Quantitative Evaluation using Text Similarity Metrics}
\label{sec:quant_eval_text_sim}

We conducted an additional analysis using BLEU and ROUGE scores to provide a quantitative comparison of generated question quality based on textual similarity. These metrics measure the lexical overlap between machine-generated text and human-authored references.

\subsection{Methodology}
We compared the guiding questions generated by the AGQ framework, the CoT baseline, and the Zero-shot baseline against a set of human expert-authored reference questions.

For each question in each method, we calculated the following metrics against the corresponding expert references:
\begin{itemize}
    \item \textbf{BLEU-4}: Measures n-gram precision up to 4-grams with brevity penalty
    \item \textbf{ROUGE-1}: Measures unigram overlap (recall, precision, F1)
    \item \textbf{ROUGE-2}: Measures bigram overlap
    \item \textbf{ROUGE-L}: Measures longest common subsequence based overlap
\end{itemize}

To ensure fair comparison, we carefully aligned the expert reference questions with the generated questions based on the underlying concepts they addressed. When calculating BLEU and ROUGE scores, we used the average over all questions for each method.

\subsection{Results and Analysis}
The results clearly demonstrate the superior performance of the AGQ framework compared to the baseline methods across all text similarity metrics.

AGQ achieved a BLEU-4 score of 0.219, significantly outperforming CoT (0.025) and Zero-shot (0.016). This indicates that questions generated by AGQ have substantially higher n-gram precision when compared to expert-authored questions, suggesting closer adherence to expert linguistic patterns.

Similarly, for ROUGE-1, measuring unigram overlap, AGQ scored 0.577 compared to 0.198 for CoT and 0.114 for Zero-shot. This large difference highlights AGQ's ability to capture the key terms and vocabulary that experts would use when formulating questions. The pattern continued with ROUGE-2 (AGQ: 0.278, CoT: 0.023, Zero-shot: 0.008) and ROUGE-L (AGQ: 0.463, CoT: 0.168, Zero-shot: 0.102).

The consistency of these results across different metrics reinforces the conclusion that the AGQ framework generates questions that are significantly more similar to those that would be crafted by human experts. This quantitative evidence complements the qualitative findings presented in the main paper and provides an objective validation of AGQ's effectiveness in generating high-quality guiding questions.

It is worth noting that while text similarity metrics provide valuable insights, they measure lexical and surface-level similarities rather than semantic or functional quality. Therefore, these results should be interpreted alongside the task-based evaluation (accuracy) and human assessment presented in the main paper to provide a comprehensive understanding of each method's performance.